%% file: main.tex
\documentclass{article}
\usepackage{ijcai17}

\usepackage{times}
\usepackage{proof}  
\usepackage{bussproofs}
\usepackage{verbatim}
\usepackage{amssymb}

\title{Strength Factors: An Uncertainty System for a Quantified Modal
  Logic} \date{\today}
\author{Naveen Sundar Govindarajulu \normalfont{and}
\textbf{Selmer Bringsjord}\\ 
Rensselaer Polytechnic Institute, Troy, NY  \\
\{naveensundarg,selmer.bringsjord\}@gmail.com}
\usepackage{url}

\usepackage{pslatex}
\usepackage{hyperref}

\usepackage{booktabs}
\usepackage{color}
\usepackage{amsfonts}
\usepackage{amsmath}
\usepackage{proof}
\usepackage[amsmath,thmmarks]{ntheorem}
\usepackage{relsize}
\usepackage{mathtools}    
\DeclarePairedDelimiterX\setc[2]{\{}{\}}{\,#1 \;\delimsize\vert\; #2\,}
\newcommand{\set}[1]{\left\lbrace #1\right\rbrace}
\DeclareMathOperator*{\argmin}{arg\,min}
\newcommand{\IGNORE}[1]{}

\newcommand{\reasonable}[2]{\ensuremath{\ \succ^{#2}_{#1}\ }}

\newcommand{\lsort}[1]{%
  \ensuremath{\mbox{\textsf{#1}}}}
\newcommand{\defsort}[2]{%
  \newcommand{#1}{\lsort{#2}}}
\defsort{\Action}{Action}
\defsort{\Time}{Time}
\defsort{\Self}{Self}

\defsort{\Agent}{Agent}
\defsort{\Entrant}{Entrant}
\defsort{\ActionType}{ActionType}
\defsort{\Moment}{Moment}
\defsort{\Boolean}{Formula}
\defsort{\PayOut}{PayOut}
\defsort{\Fluent}{Fluent}
\defsort{\Event}{Event}
\defsort{\Object}{Object}
\defsort{\RealTerm}{RealTerm}

\defsort{\Numeric}{Numeric}
\defsort{\Number}{Number}
\defsort{\Trolley}{Trolley}
\defsort{\Track}{Track}
\defsort{\Moveable}{Moveable}

\newcommand{\lsymbol}[1]{%
  \ensuremath{\mathit{#1}}}
\newcommand{\defsymbol}[2]{%
  \newcommand{#1}{\lsymbol{#2}}}
\defsymbol{\action}{action}
\defsymbol{\initially}{initially}
\defsymbol{\holds}{holds}
\defsymbol{\happens}{happens}
\defsymbol{\clipped}{clipped}
\defsymbol{\initiates}{initiates}
\defsymbol{\terminates}{terminates}
\defsymbol{\prior}{prior}
\defsymbol{\interval}{interval}

\defsymbol{\does}{does}
\defsymbol{\plans}{plans}
\defsymbol{\act}{act}
\defsymbol{\react}{react}
\defsymbol{\payTot}{pay_{tot}}
\defsymbol{\fight}{fight}
\defsymbol{\coop}{coop}
\defsymbol{\enter}{enter}
\defsymbol{\stayout}{stayout}
\defsymbol{\learns}{learns}
\defsymbol{\payoff}{payoff}
\defsymbol{\position}{position}
\defsymbol{\dead}{dead}
\defsymbol{\damaged}{damaged}

\defsymbol{\onrails}{onrails}
\defsymbol{\switch}{switch}
\defsymbol{\drop}{drop}

\newcommand{\lconstant}[1]{%
  \ensuremath{\mbox{\textsf{#1}}}}
\newcommand{\defconstant}[2]{%
  \newcommand{#1}{\lconstant{#2}}}
\defconstant{\Enter}{Enter}
\defconstant{\StayOut}{StayOut}
\defconstant{\Fight}{Fight}
\defconstant{\Acquiesce}{Acquiesce}
\defconstant{\cs}{cs }

\defconstant{\Alice}{Alice }

\defconstant{\Bob}{Bob }
\defconstant{\Gun}{gun }
\defconstant{\ta}{t_0 }
\defconstant{\tb}{t_2 }
\defconstant{\tc}{t_3 }

\newcommand{\lmodality}[1]{%
  \ensuremath{\mathbf{#1}}}
\newcommand{\defmodality}[2]{%
  \newcommand{#1}{\lmodality{#2}}}
\defmodality{\common}{C}
\defmodality{\knows}{K}
\defmodality{\believes}{B}
\defmodality{\withholds}{W}

\defmodality{\perceives}{P}
\defmodality{\mental}{M}

\defmodality{\desires}{D}
\defmodality{\intends}{I}
\defmodality{\says}{S}
\defmodality{\ought}{O}

\newcommand{\sep}{\ \big\lvert \ }

\usepackage{booktabs}
\usepackage{mdframed}
\usepackage{paralist}

\usepackage{color}

\usepackage[usenames,dvipsnames,svgnames,table]{xcolor}

\newcommand{\DCEC}{\ensuremath{{\mathcal{{DCEC}}}}}
\newcommand{\mC}{\ensuremath{{\mathcal{{\mu C}}}}}

\newcommand{\type}[1]{\textsf{#1}}

\begin{document}
\date{March 2017}
\maketitle

\begin{abstract}
  We present a new system $\mathcal{S}$ for handling uncertainty in a
  quantified modal logic (first-order modal logic). The system is
  based on both probability theory and proof theory and is derived
  from Chisholm's epistemology. We concretize Chisholm's system by
  grounding his undefined and primitive (i.e. foundational) concept of
  \textbf{reasonableness} in probability and proof theory. We discuss
  applications of the system. The system described below is a work in
  progress; hence we end by presenting a list of future challenges.

\end{abstract}


\section{Introduction}
\label{sect:intro}
We introduce a new system $\mathcal{S}$ for talking about uncertainty
of iterated beliefs in a quantified modal logic with belief
operators. The quantified modal logic we use is based on the
\textbf{deontic cognitive event calculus} (\DCEC), which belongs to
the family of \textbf{cognitive calculi} that have been used in
modeling complex cognition. Here, we use a subset of $\DCEC$ that we
term \textbf{micro cognitive calculus} ($\mC$). Specifically, we add a
system of uncertainty derived from Chisholm's epistemology
\cite{theory.of.knowledge3.chisholm}.\footnote{See the SEP entry on
  Chisholm for a quick overview of Chisholm's epistemology:
  \url{https://plato.stanford.edu/entries/chisholm/\#EpiIEpiTerPriFou}.}
The system $\mathcal{S}$ is a work in progress and hence the
presentation here will be abstract in nature.

One of our primary motivations is to design a system of uncertainty
that is easy to use in end-user facing systems. There have been many
studies that show that laypeople have difficulty understanding raw
probability values (e.g. see \cite{kaye_1991}); and we believe that
our approach borrowed from philosophy can pave the way for systems
that can present uncertain statements in a more understandable format
to lay users.

$\mathcal{S}$ can be useful in systems that have to interact with
humans and provide justifications for their uncertainty. As a
demonstration of the system, we apply the system to provide a solution
to the lottery paradox. Another advantage of the system is that it can
be used to provide uncertainty values for counterfactual
statements. Counterfactuals are statements that an agent knows for
sure are false. Among other cases, counterfactuals are useful when
systems have to explain their actions to users (\emph{If I had not
  done $\alpha$, then $\phi$ would have happened}).  Uncertainties for
counterfactuals fall out naturally from our system. Before we discuss
the calculus and present $\mathcal{S}$, we go through relevant prior
work.



\section{Prior Work}

Members in the family of cognitive calculi have been used to formalize
and automate highly intensional reasoning processes.\footnote{By
  ``intensional processes'', we roughly mean processes that take into
  account knowledge, beliefs, desires, intentions, etc. of
  agents. Compare with extensional systems such as first-order logic
  that do not take into account states of minds of other agents. This
  is not be confused with ``intentional'' systems which would be
  modeled with intensional systems. See \cite{zalta1988intensional}
  for a detailed treatment of intensionality.} More recently, using
\DCEC\ we have presented an automation of \textbf{the doctrine of
  double effect} in \cite{nsg_sb_dde_2017}.\footnote{This work will be
  presented at IJCAI 2017.}  We quickly give an overview of the
doctrine to illustrate the scope and expressivity of cognitive calculi
such as \DCEC. The doctrine of double effect is an ethical principle
that has been shown to be used by both untrained laypeople and experts
when faced with moral dilemmas; and it plays a central role in many
legal systems. Moral dilemmas are situations in which all available
options have both good and bad consequences. The doctrine states that
an action $\alpha$ in such a situation is permissible \emph{iff}
--- \begin{inparaenum}[(1)] \item it is morally neutral; \item the net
  good consequences outweigh the bad consequences by some large amount
  $\gamma$; and \item at least one or more of the good consequences
  are \emph{intended}, and none of the bad consequences are
  intended. \end{inparaenum} The conditions require both intensional
operators and a calculus (e.g. the event calculus) for modeling
commonsense reasoning and the physical world. Other tasks automated by
cognitive calculi include the false-belief task
\cite{ArkoudasAndBringsjord2008Pricai} and \textit{akrasia}
(succumbing to temptation to violate moral principles)
\cite{akratic_robots_ieee_n}.\footnote{Arkoudas and Bringsjord
  \shortcite{ArkoudasAndBringsjord2008Pricai} introduced the general
  family of \textbf{cognitive event calculi} to which \DCEC\ belongs.}
Each cognitive calculus is a sorted (i.e.\ typed) quantified modal
logic (also known as sorted first-order modal logic).  Each calculus
has a well-defined syntax and proof calculus.  The proof calculus is
based on natural deduction
\cite{gentzen_investigations_into_logical_deduction}, and includes all
the introduction and elimination rules for first-order logic, as well
as inference schemata for the modal operators and related structures.

On the uncertainty and probability front, there have been many logics
of probability, see \cite{sep-logic-probability} for an
overview. Since our system builds upon probabilities, our approach
could use a variety of such systems. There has been very little work
in uncertainty systems for first-order modal logics. Among first-order
systems, the seminal work in \cite{halpern1990analysis} presents a
first-order logic with modified semantics to handle probabilistic
statements. We can use such a system as the foundation for our work,
and use it to define the base probability function $\mathbf{Pr}$ used
below. (Note that we leave $\mathbf{Pr}$ unspecified for now.)

\section{The Formal System} 

The formal system $\mC$ is a modal extension of the the event calculus. The
event calculus is a multi-sorted first-order logic with a family of
axiom sets. The exact axiom set is not important. The primary sorts in
the system are shown below.  

\begin{scriptsize}
\begin{tabular}{lp{5.89cm}}  
\toprule
\textbf{Sort}    & \textbf{Description} \\
\midrule
\type{Agent} & Human and non-human actors.  \\

\type{Moment} or  \type{Time} &  Time points and intervals. E.g. simple, such as $t_i$, or complex, such as
$birthday(son(jack))$. \\

 \type{Event} & Used for events in the domain. \\
 \type{ActionType} & Action types are abstract actions.  They are
  instantiated at particular times by actors.  E.g.: ``eating''
                     vs. ``jack eats.''\\
 \type{Action} & A subtype of \type{Event} for events that occur
  as actions by agents. \\
 \type{Fluent} & Used for representing states of the world in the
  event calculus. \\
\bottomrule
\end{tabular}
\end{scriptsize}

Full \DCEC\ has a suite of modal operators and inference
schemata. Here we focus on just two: an operator for belief
$\believes$ and an operator for perception $\perceives$. The syntax of
and inference schemata of the system are shown below. $\mathit{S}$ is
the set of all sorts, $\mathit{f}$ are the core function symbols, $t$
shows the set of terms, and $\phi$ is the syntax for the formulae.

\begin{scriptsize}
\begin{mdframed}[linecolor=white, frametitle=Syntax, frametitlebackgroundcolor=gray!25, backgroundcolor=gray!10, roundcorner=8pt]
\begin{equation*}
 \begin{aligned}
    \mathit{S} &::= 
    \left\{\begin{aligned}
      & \Object \sep \Agent \sep \Self \sqsubset \Agent \sep \ActionType \sep \Action \sqsubseteq
      \Event \sep \\ &\Moment \sep \Boolean \sep \Fluent \sep \Numeric\\
    \end{aligned} \right.
    \\ 
    \mathit{f} &::= \left\{
    \begin{aligned}
      & \action: \Agent \times \ActionType \rightarrow \Action \\
      &  \initially: \Fluent \rightarrow \Boolean\\
      &  \holds: \Fluent \times \Moment \rightarrow \Boolean \\
      & \happens: \Event \times \Moment \rightarrow \Boolean \\
      & \clipped: \Moment \times \Fluent \times \Moment \rightarrow \Boolean \\
      & \initiates: \Event \times \Fluent \times \Moment \rightarrow \Boolean\\
      & \terminates: \Event \times \Fluent \times \Moment \rightarrow \Boolean \\
      & \prior: \Moment \times \Moment \rightarrow \Boolean\\
    \end{aligned}\right.\\
        \mathit{t} &::=
    \begin{aligned}
      \mathit{x : S} \sep \mathit{c : S} \sep f(t_1,\ldots,t_n)
    \end{aligned}
    \\ 
    \mathit{\phi}&::= \left\{ 
    \begin{aligned}
     t: &\Boolean \sep  \neg \phi \sep \phi \land \psi \sep \phi \lor
     \psi \sep \phi \rightarrow \psi \sep \phi \leftrightarrow \psi\\
     & \perceives (a,t,\phi)  \sep \believes(a,t,\phi) 
      \end{aligned}\right.
  \end{aligned}
\end{equation*}
\end{mdframed}
\end{scriptsize}
\vspace{0.08in}

The above calculus lets us formalize statements of the form \emph{``John
believes now that Mary perceived that it was raining.''} One formalization could be: 
\begin{footnotesize}
\begin{equation*}
 \begin{aligned}
\exists t<now: \believes\Big(\mathit{john}, \mathit{now},
\perceives\big(\mathit{mary}, t, \holds(raining, t)\big)\Big)  
  \end{aligned}
\end{equation*}
\end{footnotesize}
The figure below shows the inference schemata for
\mC.  $R_\mathbf{P}$ captures that perceptions get turned into
beliefs. $R_\mathbf{B}$ is an inference schema that lets us model
idealized agents that have their beliefs closed under the \mC\ proof
theory.  While normal humans are not deductively closed, this lets us
model more closely how deliberate agents such as organizations and
more strategic actors reason. Assume that there is a background set of
axioms $\Gamma$ we are working with.

\begin{footnotesize}
\begin{mdframed}[linecolor=white, frametitle=Inference Schemata, frametitlebackgroundcolor=gray!25, backgroundcolor=gray!10, roundcorner=8pt]
\begin{equation*}
\begin{aligned}
&\hspace{62pt}  \infer[{[R_{\perceives}]}]{\believes(a,t_2,\phi)}{\perceives(a,t_1,\phi_1),
    \ \ \Gamma \vdash t_1 < t_2} \\
\vspace{20pt}
  &\infer[{[R_{\believes}]}]{\believes(a,t,\phi)}{\believes (a,t_1,\phi_1),
    \ldots, \believes (a,t_m,\phi_m), \ 
    \  \{\phi_1, \ldots, \phi_m\}\vdash  \phi, \ \ \Gamma\vdash t_i < t}  \\
\end{aligned}
\end{equation*}
\end{mdframed}
\end{footnotesize}

\section{The Uncertainty System $\mathcal{S}$}
In the uncertainty system, we augment the belief modal operators with
a discrete set of uncertainty factors termed as \emph{strength
  factors}. The factors are not arbitrary and are based on how
derivable a proposition is for a given agent.

Chisholm's epistemology has a primitive undefined binary relation that
he terms \textbf{reasonableness} with which he defines a scale of
strengths for beliefs one might have in a proposition. Note that
Chisholm's system is agent free while ours is agent-based. Let
$\phi\ \mathcal{\succ}^a_t\ \psi$ denote that $\phi$ is more
reasonable than $\psi$ to an agent $a$ at time $t$. We require that
$\reasonable{t}{a}$ be \emph{asymmetric}: i.e., \emph{irreflexive} and
\emph{anti-symmetric}.  That is, for all $\phi$,
$\phi \not \reasonable{t}{a} \phi$; and for all $\phi$ and $\psi$,

\begin{footnotesize}
\begin{equation*}
 \begin{aligned}
\left(\phi \reasonable{t}{a} \psi \right) \Rightarrow \left(\psi \not\reasonable{t}{a}
\phi\right)  
 \end{aligned}
\end{equation*}
\end{footnotesize}

We also require that $ \reasonable{t}{a} $ be transitive. In addition
to these conditions, we have the following five requirements governing
how $\reasonable{t}{a}$ interacts with the logical connectives
$\land, \lnot$ and $\mathbf{B}$ (the first three conditions can
be derived from the defintion of $\succ$ sketched out later):
\begin{footnotesize}
\begin{equation*}
 \begin{aligned}
&[\mathbf{C}_{\land_1}] \ 
\big(\psi_1\reasonable{t}{a} \phi_1\big)  \mbox{ and } \big(\psi_2
\reasonable{t}{a} \phi_2\big) \Rightarrow\Big(\psi_1 \reasonable{t}{a} \phi_1 \land \phi_2 \Big)\\
&[\mathbf{C}_{\land_2}] \ 
\big(\psi_1 \land \psi_2 \reasonable{t}{a} \phi\big)
\Rightarrow\Big[\big(\psi_1 \reasonable{t}{a} \phi\big)\mbox{and }
\big(\psi_2 \reasonable{t}{a} \phi \big)\Big]\\
&[\mathbf{C}_{\lnot}] \ \mbox{There is no }\phi \mbox{ such that } \left(\bot
\reasonable{t}{a}\phi\right) \mbox{; and for all }\phi \left(\phi
\reasonable{t}{a}\bot\right)\\
&[\mathbf{C}_{\mathbf{B}_1}] \ \Big( \believes(a,t,\phi) \reasonable{t}{a}
\believes(a,t,\lnot \phi) \Big)\Rightarrow  \Big(\believes(a,t,\phi) \reasonable{t}{a}
\lnot \believes(a,t, \phi)\Big)\\
&[\mathbf{C}_{\mathbf{B}_2}] \ \mbox{For all } \phi, \ \left[\begin{aligned} &
  \big(\believes(a,t,\phi) \reasonable{t}{a} \believes(a,t,\lnot\phi)
\big)\mbox{ or }\\
   &\big(\believes(a,t,\lnot \phi) \reasonable{t}{a}
   \believes(a,t,\phi) \big)
 \end{aligned}\right]
 \end{aligned}
\end{equation*}
\end{footnotesize}

We also add a \textbf{belief consistency condition} which requires
that:

\begin{footnotesize}
\begin{equation*}
  \begin{aligned}
\Big(\Gamma\vdash\mathbf{B}^p(a,t,\phi)\Big)\Leftrightarrow \Big(\Gamma\not\vdash\mathbf{B}^p(a,t,\lnot\phi)\Big)
  \end{aligned}
\end{equation*}
\end{footnotesize}

For convenience, we define a new operator, the withholding operator
$\withholds$ (this is simply syntactic sugar):

\begin{footnotesize}
\begin{equation*}
 \begin{aligned}
\withholds(a,t,\phi) \equiv \lnot\believes(a,t,\phi) \land \lnot\believes(a,t,\lnot \phi) 
  \end{aligned}
\end{equation*}
\end{footnotesize}

We now reproduce Chisholm's system below. Note the formula used in the
definitions below are meta-formula and not strictly in \mC. 
\begin{small}
\begin{mdframed}[linecolor=white, frametitle=Strength Factor Definitions, frametitlebackgroundcolor=gray!25, backgroundcolor=gray!10, roundcorner=8pt]

\begin{description}

\item[Acceptable] An agent $a$ at time $t$ finds $\phi$ acceptable
  \emph{iff} withholding $\phi$ is not more reasonable than believing
  in $\phi$.

\begin{footnotesize}
\begin{equation*}
  \begin{aligned}
\believes^{\mathsf{1}}(a,t,\phi) \Leftrightarrow \left\{ \begin{aligned} 
   & \withholds(a,t, \phi) \not \succ^{a}_{t}\ \believes(a,t,\phi)
    \mbox{; or}\\ 
    &\Big(\lnot \believes(a,t,\phi)
    \land \lnot \believes(a,t,\lnot\phi)\Big)\not\succ^{a}_{t}\  \believes(a,t,\phi)\end{aligned}\right.
  \end{aligned}
\end{equation*}
\end{footnotesize}

\item[Some Presumption in Favor] An agent $a$ at time $t$ has some
  presumption in favor of $\phi$ \emph{iff} believing $\phi$ at $t$ is more reasonable than
  believing $\lnot\phi$ at  time $t$:

\begin{footnotesize}
\begin{equation*}
  \begin{aligned}
\believes^{\mathsf{2}}(a,t,\phi) \Leftrightarrow \
    \Big(\believes(a,t,\phi)&\reasonable{t}{a} \believes (a,t, \lnot \phi)\Big)
  \end{aligned}
\end{equation*}
\end{footnotesize}

\item[Beyond Reasonable Doubt] An agent $a$ at time $t$ has beyond
  reasonable doubt in $\phi$ \emph{iff} believing $\phi$ at $t$ is more reasonable than
  withholding $\phi$ at  time $t$:

\begin{footnotesize}
\begin{equation*}
  \begin{aligned}
\believes^{\mathsf{3}}(a,t,\phi) \Leftrightarrow \left\{ \begin{aligned} 
    \believes(a,t,\phi)&\reasonable{t}{a} \withholds(a,t, \phi)
    \mbox{; or}\\ 
    \Big(\believes(a,t,\phi)&\succ^{a}_{t}\Big( \lnot \believes(a,t,\phi)
    \land \lnot \believes(a,t,\lnot\phi)\Big)\end{aligned}\right.
  \end{aligned}
\end{equation*}
\end{footnotesize}

\item[Evident] A formula $\phi$ is evident to  an agent $a$ at time $t$
  \emph{iff} $\phi$ is beyond reasonable doubt and if there is a $\psi$
  such that believing $\psi$ is more reasonable for $a$ at time $t$
  than believing $\phi$, then $a$ is certain about $\psi$ at time $t$.

\begin{footnotesize}
\begin{equation*}
  \begin{aligned}
\believes^{\mathsf{4}}(a,t,\phi) \Leftrightarrow \left\{ \begin{aligned} 
 &  \believes^{\mathsf{3}}(a,t,\phi) 
    \land \\ 
     &\exists \psi:\left[\begin{aligned} \believes(a,t,\psi) & \succ^a_t
     \believes(a,t,\phi)\\
\Rightarrow &\believes^{\mathsf{5}}(a,t,\psi)\end{aligned}\right]\end{aligned}\right.
  \end{aligned}
\end{equation*}
\end{footnotesize}

\item[Certain] An agent $a$ at time $t$ is certain about $\phi$
  \emph{iff} $\phi$ is beyond reasonable doubt and there is no $\psi$
  such that believing $\psi$ is more reasonable for $a$ at time $t$
  than believing $\phi$.

\begin{footnotesize}
\begin{equation*}
  \begin{aligned}
\believes^{\mathsf{5}}(a,t,\phi) \Leftrightarrow \left\{ \begin{aligned} 
   \believes^{\mathsf{3}}(a,t,\phi) & 
    \land \\ 
    \lnot \exists \psi: \believes(a,t,\psi) & \succ^a_t \believes(a,t,\phi)\end{aligned}\right.
  \end{aligned}
\end{equation*}
\end{footnotesize}

\end{description} 
\end{mdframed}
\end{small}
\vspace{0.1in}

The above definitions are from Chisholm but more rigorously formalized
in \mC. The definitions and the conditions
$\{[\mathbf{C}_{\land_1}], [\mathbf{C}_{\land_2},
[\mathbf{C}_{\lnot}], [\mathbf{C}_{\mathbf{B}_1}],
[\mathbf{C}_{\mathbf{B}_1}]\}$ give us the following theorem.

\begin{small}
\begin{mdframed}[linecolor=white, frametitle= {Theorem}: Higher Strength subsumes
  Lower Strength, frametitlebackgroundcolor=gray!25,  backgroundcolor=gray!10, roundcorner=8pt]
For any $p$ and $q$, if $p>q$, we have: $\believes^p(a,t,\phi) \Rightarrow \believes^q(a,t,\phi) $
\end{mdframed}
\end{small}
\vspace{0.1in}

\noindent{\textbf{Proof}}: $\mathbf{B}^5\Rightarrow \mathbf{B}^3$ and
$\mathbf{B}^4 \Rightarrow \mathbf{B}^3$ by definition.
$\mathbf{B}^5 \Rightarrow \mathbf{B}^4$ by the second clause in the
definitions of $\mathbf{B}^4$ and $\mathbf{B}^5$.
$\mathbf{B}^3 \Rightarrow \mathbf{B}^1$ by the asymmetry property of $\reasonable{t}{a}$.

For $\mathbf{B}^2 \Rightarrow \mathbf{B}^1$, we have a proof by
contradiction. Assume that (in shorthand):
$$\left(\mathbf{B}\phi \succ \mathbf{B}\lnot\phi \right) \mbox{ but} 
\left(\lnot\mathbf{B}\phi \land \lnot \mathbf{B}\lnot \phi\right)
\succ \mathbf{B}\phi$$

Using $[\mathbf{C}_{\mathbf{B}_1}]$ on the former and  $[\mathbf{C}_{\land_2}]$ on the latter, we get

$$\mathbf{B}\phi \succ \lnot \mathbf{B}\phi \mbox{ and }   \lnot
  \mathbf{B} \phi\succ \mathbf{B}\phi $$

Using transitivity, we get $ \mathbf{B} \phi\succ  
\mathbf{B}\phi $. This violates irreflexivity, therefore $\mathbf{B}^2 \Rightarrow \mathbf{B}^1$.

For $\mathbf{B}^3 \Rightarrow \mathbf{B}^2$, if the condition for
$\mathbf{B}^2$ does not hold, by $\mathbf{C}_{\mathbf{B}_2}$ we have:

$$\mathbf{B}\lnot\phi \succ \mathbf{B}\phi$$

Using the condition for $\mathbf{B}^3$ and transitivity, we get 

$$\mathbf{B}\lnot\phi \succ \lnot \mathbf{B}\phi \land \lnot
\mathbf{B}\lnot\phi$$ giving us $\mathbf{B}^3\lnot\phi$, and we
started with $\mathbf{B}^3\phi$.  This violates the belief consistency condition. $\blacksquare$

 The definitions almost give us $\mathcal{S}$ except for the
fact that $\succ^a_t$ is undefined. While Chisholm gives a careful and
informal analysis of the relation, he does not provide a more precise
definition. Such a definition is needed for automation.  We provide a
three clause defintion that is based on both probabilities and proof
theory.

There are many probability logics that allow us to define
probabilities over formulae. They are well studied and understood for
propositional and first-order logics. Let $\mathcal{L}$ be the set of
all formulae in \mC. Let $\mathcal{L}_p$ be a pure first-order
subset of $\mathcal{L}$.  Assume that we have the following
partial probability function defined over $\mathcal{L}_p$\footnote{Something
similar to the system in \cite{halpern1990analysis} that accounts for
probabilities as statistical information or degrees of belief can
work.}:
\begin{footnotesize}
\begin{equation*}
  \begin{aligned} \mathbf{Pr}: \Agent \times \Moment \times \Boolean \mapsto
\mathbb{R}\end{aligned}
\end{equation*}
\end{footnotesize}\vspace{-0.05in} Then we have the first clause of our definition for $\succ^a_t$.
\begin{mdframed}[linecolor=white, frametitle=Clause I. Defining $\succ$, frametitlebackgroundcolor=gray!25, backgroundcolor=gray!10, roundcorner=8pt]
\begin{footnotesize}
If $\mathbf{Pr} (a,t, \phi)$ and $\mathbf{Pr}(a,t,\psi)$ are defined then:
\begin{equation*}
  \begin{aligned}
\Big(\phi  \reasonable{t}{a}  \psi\Big)  \Leftrightarrow  \Big(\mathbf{Pr}(a,t, \phi) >
\mathbf{Pr}(a,t, \psi)\Big)
  \end{aligned}
\end{equation*}
\end{footnotesize}
\vspace{-0.2in}
\end{mdframed}

\vspace{0.1in} We might not always have meaningful probabilities for
all propositions. For example, consider propositions of the form
\emph{``I believe that Jack believes that $\phi$.''} It is hard to get
precise numbers for such statements. In such situations, we might look
at the ease of derivation of such statements given a knowledge base
$\Gamma$. \footnote{Another possible mechanism can leverage
  Dempster-Shafer models of uncertainty for first-order
  logic \cite{nunezetal13folinds}.} Given two competing statements
$\phi$ and $\psi$, we can say one is more reasonable than the other if
we can easily derive one more than the other from $\Gamma$. This
assumes that we can derive $\phi$ and $\psi$ from $\Gamma$. We assume
we have a cost function $\rho: \mathsf{Proof} \mapsto \mathbb{R}^+$
that lets us compute costs of proofs.  There are many ways of
specifying such functions. Possible candidates are length of the
proof, time for computing the proof, depth vs breadth of the proof,
unique symbols used in the proof etc. We leave this choice unspecified
but any such function could work here. Let $\vdash_{a,t}$ denote
provability w.r.t. to agent $a$ at time $t$.

\begin{mdframed}[linecolor=white, frametitle=Clause II. Defining $\succ$, frametitlebackgroundcolor=gray!25, backgroundcolor=gray!10, roundcorner=8pt]
\begin{footnotesize}

If one of $\mathbf{Pr} (a,t, \phi)$ and $\mathbf{Pr}(a,t,\psi)$ is not
defined, but if $\Gamma\vdash_{a,t}\phi$ and $\Gamma\vdash_{a,t}
\psi$:

\begin{equation*}
  \begin{aligned}
\Big(\phi \reasonable{t}{a} \psi\Big)  \Leftrightarrow
\Big(\rho\big(\Gamma\vdash_{a,t}\phi\big) < \rho\big(\Gamma\vdash_{a,t}\psi\big)\Big)
  \end{aligned}
\end{equation*}
\end{footnotesize}
\vspace{-0.15in}

\end{mdframed}

\vspace{0.05in}

Clauses I and II might not always be applicable as the premises in the
definitions might not always hold. A more common case could be when
we cannot derive the propositions of interest from our background set
of axioms $\Gamma$. For example, if we are interested in the
uncertainty values for statements that we know are false, then it
should be the case that they be not derivable from our background set
of axioms. In this situation, we look at $\Gamma$ and see what minimal
changes we can make to it to let us derive the proposition of
interest. Trivially, if we cannot derive $\phi$ from $\Gamma$, we can
add it to $\Gamma$ to derive it, as $\Gamma + {\phi} \vdash
\phi$. This is not desirable for two reasons.

First, simply adding to $\Gamma$ might result in a contradiction. In
such cases we would be looking at removing a minimal set of
statements $\Lambda$ from $\Gamma$. Second, we might prefer to add a
more simpler set of propositions $\Theta$ to $\Gamma$ rather than
$\phi$ itself to derive $\phi$. Recapping, we go from $(1)$ to $(2)$
below:

\begin{footnotesize}
\begin{equation*}
  \begin{aligned}
\Gamma &\not \vdash \phi \ \  (1)\\
 \Gamma \cup \Theta - \Lambda  &\vdash \phi  \ \ (2)\\
  \end{aligned}
\end{equation*}
\end{footnotesize}

When we go from $(1)$ to $(2)$ we would like to modify the background
axioms as minimally as possible.  Assume that we have a similarity
 function $\pi$ for sets of formulae.  We then choose $\Theta$ and
 $\Lambda$ as given below ($\mathit{Con}[S]$ denotes that $S$ is consistent):
 \begin{footnotesize}
 \begin{equation*}    \begin{aligned}
 \langle \Theta, \Lambda \rangle = \argmin_{\langle \Theta,
  \Lambda\rangle}\  &\pi(\Gamma, \Gamma \cup \Theta - \Lambda); \mbox{ such
   that }\left\{\begin{aligned} &\Gamma \cup \Theta - \Lambda \vdash \phi; \mbox{ and} \\
&\mathit{Con}\big[\Gamma \cup \Theta -\Lambda\big]
   \end{aligned}\right.\end{aligned}
 \end{equation*}
\end{footnotesize}

Consider a statement such as \emph{``It rained last week''} when it
did not actually rain last week, and another statement such as \emph{``The moon
  is made of cheese.''} Both statements denote things that did not
happen, but intuitively it seems that former should be more easier to
accept from what we know than the latter. There are many similarity
measures which can help convey this. Analogical reasoning is one such
possible measure of similarity. If the new formulae are structurally
similar to existing formulae, then we might be more justified in
accepting such formulae. For example, one such measure could be the
analogical measure used by us in \cite{GIfromLP_ijcai13_short}.

Now we have the formal mechanism in place for defining the final
clause in our definition for our reasonableness. Let
$\delta^a_t(\Gamma, \phi)$ be the distance between $\Gamma$ and closest
consistent set under $\pi$ that lets us prove $\phi$:

\begin{footnotesize}
\begin{equation*}
  \begin{aligned}
\delta^a_t(\Gamma, \phi) \equiv \min_{\langle \Theta, \Lambda\rangle}\set{\pi\big(\Gamma, \Gamma
\cup \Theta-\Lambda \big)   \,\middle|\ \begin{aligned} &\big(\Gamma \cup \Theta -\Lambda\big)
\vdash^a_t \psi \mbox{; and} \\ &\mathit{Con}\big[\Gamma \cup \Theta -\Lambda\big]\end{aligned}}
  \end{aligned}
\end{equation*}
\end{footnotesize}

\begin{mdframed}[linecolor=white, frametitle= Clause III. Defining $\succ$, frametitlebackgroundcolor=gray!25, backgroundcolor=gray!10, roundcorner=8pt]
\begin{footnotesize}

If one of $\mathbf{Pr} (a,t, \phi)$ and $\mathbf{Pr}(a,t,\psi)$ is not
defined, and one of $\Gamma\vdash_{a,t}\phi$ and $\Gamma\vdash_{a,t}
\psi$ does not hold, then 

\begin{equation*}
  \begin{aligned}
\Big(\phi \reasonable{t}{a} \psi\Big)  \Leftrightarrow 
\Big[\delta^a_t(\Gamma, \phi) < \delta^a_t(\Gamma, \psi) \Big]
  \end{aligned}
\end{equation*}
\end{footnotesize}

\end{mdframed}

\vspace{0.1in}

The final piece of $\mathcal{S}$ is inference rules for belief
propagation with uncertainty values. This is quite
straightforward. Inferences propagate uncertainty values from the
premises with the lowest strength factor; and inferences happen only
with beliefs that are close in their uncertainty values, with maximum
difference being parametrized by $u$, with default $u = 2$.

\begin{footnotesize}
\begin{mdframed}[linecolor=white, frametitle=Inference Schemata for $\mathcal{S}$, frametitlebackgroundcolor=gray!25, backgroundcolor=gray!10, roundcorner=8pt]
\begin{equation*}
\begin{aligned}
\vspace{20pt}
&\hspace{45pt}  \infer[{[R^s_{\perceives}]}]{\believes^5(a,t_2,\phi)}{\perceives(a,t_1,\phi_1),
    \ \ \Gamma \vdash t_1 < t_2} \\
\vspace{0.9in}
\\
 & \hspace{-5pt}\infer[{[R^s_{\believes}]}]{\believes^{min(s_1, \ldots, s_m)}(a,t,\phi)}{\believes^{s_1}(a,t_1,\phi_1),
    \ldots, \believes^{s_m}(a,t_m,\phi_m), \{\phi_1, \ldots,
    \phi_m\}\vdash  \phi, \Gamma\vdash t_i < t}  \\
& \ \ \ \mbox{with  }\ \ max(\{s_1,\ldots, s_m\}) - min(\{s_1,\ldots, s_m\}) \leq u
\end{aligned}
\end{equation*}
\end{mdframed}
\end{footnotesize}


\section{Usage}

In this section, we illustrate $\mathcal{S}$ by applying it solve
problems of foundational interest such as the lottery paradox
\cite[p.~197]{kyburg1961probability} and a toy version of a more real
life example, a murder mystery example (following in the traditions of
logic pedagogy). Finally, we very briefly sketch abstract scenarios in
which $\mathcal{S}$ can be used to generate uncertainty values for
counterfactual statements and to generate explanations for actions.

\subsection{Paradoxes: Lottery Paradox}
In the lottery paradox, we have a situation in which an agent $a$
comes to believe $\phi$ and $\lnot \phi$ from a seemingly consistent
set of premises $\Gamma_L$ describing a lottery. Our solution to the
paradox is that the agent simply has different strengths of beliefs in
the proposition and its negation. We first go over the paradox
formalized in $\mC$ and then present the solution.

\input{lottery_paradox}
\vspace{-0.06in}

\subsection{Application: Solving a Murder}
We look at a toy example in which an agent $\mathsf{s}$ has to solve a
murder that happened at time $t_3$. $\mathsf{s}$ believes that either
$\mathsf{Alice}$ or $\mathsf{Bob}$ is the murderer. The agent knows
that there is a gun \textsf{gun} involved in the murder and that the
owner of the gun at $t_3$ committed the murder.  $s$ also knows that
$\mathsf{Alice}$ is the owner of the gun initially at time $t_0$.

\begin{small}
\begin{mdframed}[linecolor=white, frametitle= Presumption in Favor of
  $\mathsf{Alice}$ Being the Murderer, frametitlebackgroundcolor=gray!25, backgroundcolor=gray!10, roundcorner=8pt]
  From just these facts, the agent has some presumption for believing
  that $\mathsf{Alice}$ is the murderer.
\end{mdframed}
\end{small}
\vspace{0.1in}

\noindent{\textbf{Proof Sketch}}: All the above statements can be
taken as certain beliefs $\believes^5$ of $s$. For convenience, we consider the
formulae directly without the belief operators.

\begin{figure}[h!]
 \begin{minipage}[b]{\linewidth}
  \caption{$\mathsf{Alice}$ is not the murder: $\believes\Big(s, t, \lnot \mathit{Murderer}(\mathsf{Alice})\Big)$}
  \medskip
  \label{fig:proof2}
  {\includegraphics[scale=0.31]{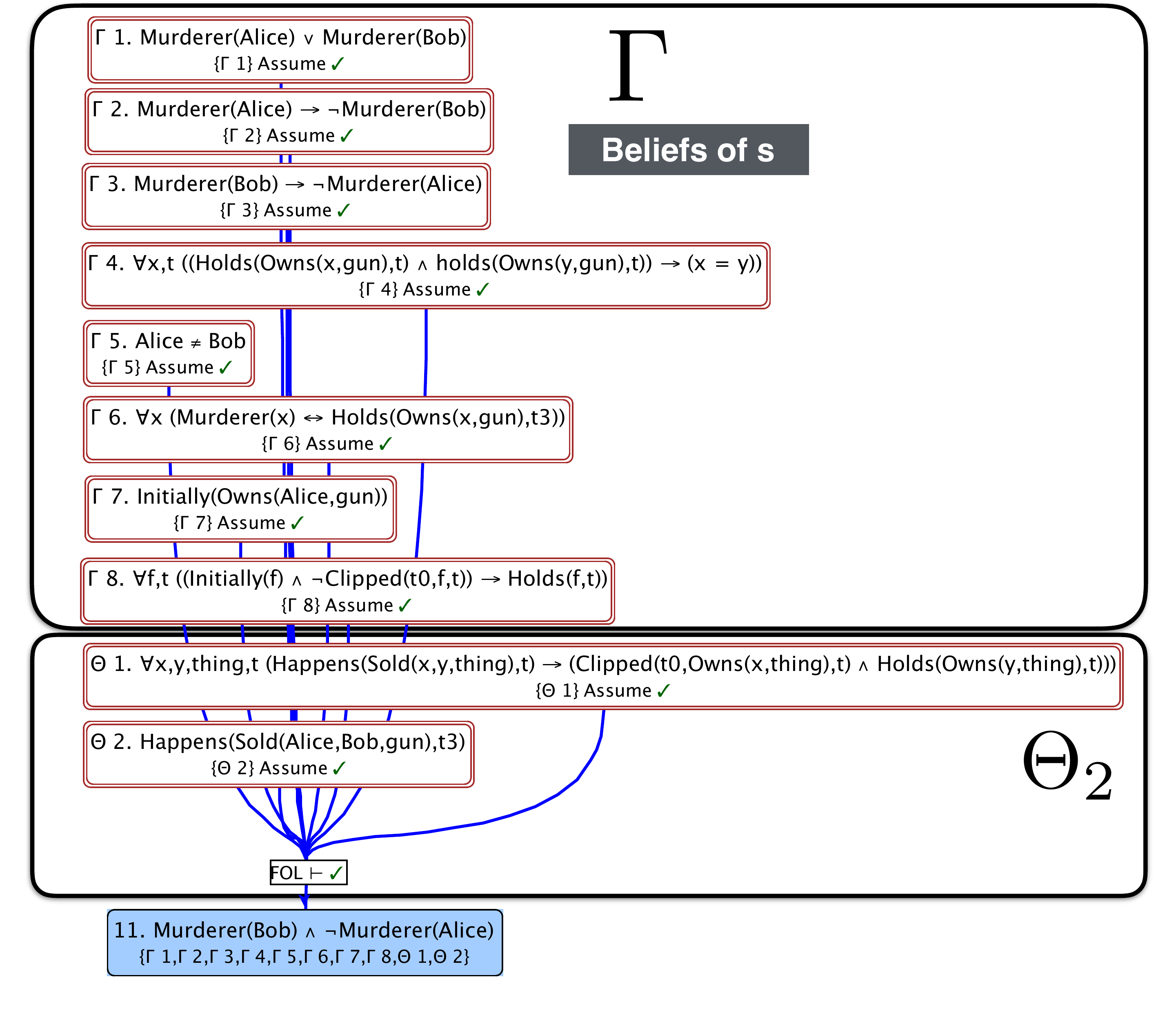}}
 \end{minipage}
\end{figure}

In order to prove the above, we need to prove that it is easier for the
agent to derive that $\mathsf{Alice}$ is the murderer than to derive
that $\mathsf{Alice}$ is not the murderer. First, to prove the former, the
agent just has to assume that $\mathsf{Alice}$'s ownership of the gun
did not change from $t_0$ to $t_3$. Second, in order for the agent to believe
that $\mathsf{Alice}$ did not commit the murder but $\mathsf{Bob}$
committed it, the agent must be willing to admit that something
happened to change $\mathsf{Alice}$'s ownership of the gun from time
$t_0$ to $t_3$ that results in $\mathsf{Bob}$ owning the gun. One
possibility is that $\mathsf{Alice}$ simply sold the gun to
$\mathsf{Bob}$. Both the scenarios are shown as proofs in the Slate
theorem proving workspace \cite{Slate_at_CMNA08} in the
Appendix. Figure~\ref{fig:proof1} shows a proof modulo belief
operators of $\believes(s, now, \mathit{Murderer}(\mathsf{Alice}))$
from $\Gamma \cup \Theta_1$ and Figure~\ref{fig:proof2} shows a proof
of $\believes(s, now, \lnot \mathit{Murderer}(\mathsf{Alice}))$ from
$\Gamma \cup \Theta_2$. 

If we assume that $\Theta_1$ and $\Theta_2$ exhaust the space of
allowed additions, then it easy to see how syntactic measures of
complexity will yield that
$\delta^a_t\big(\Gamma, \Gamma\cup\Theta_1\big) < \delta^a_t\big(\Gamma,
\Gamma\cup\Theta_2\big)$ as $\Theta_2$ is more complex than
$\Theta_1$. This lets us derive that $s$ has some presumption in
favor of $\mathit{Murderer}(\mathsf{Alice})$. $\blacksquare$

What happens if the agent knows or has a belief with certainty that
$\mathsf{Alice}$'s ownership of the gun did not change from $t_0$ to
$t_3$? 

\begin{small}
\begin{mdframed}[linecolor=white, frametitle= Beyond Reasonable Doubt
  that 
  $\mathsf{Alice}$ is the Murderer, frametitlebackgroundcolor=gray!25, backgroundcolor=gray!10, roundcorner=8pt]
  If the agent is certain that $\mathsf{Alice}$'s ownership of the gun
  did not change from $t_0$ till $t_3$, the agent has beyond
  reasonable doubt that she is the murderer.
\end{mdframed}
\end{small}
\vspace{0.05in}

\noindent \textbf{Proof Sketch:} In this case we directly have that:\begin{footnotesize}
\begin{equation*}
  \begin{aligned}
\Gamma &\vdash\believes(s, now, \mathit{Murderer}(\mathsf{Alice}))\\
\Gamma \not &\vdash \lnot \believes(s, now,
\mathit{Murderer}(\mathsf{Alice}))\\
\Gamma \not &\vdash \lnot \believes(s, now, \lnot
\mathit{Murderer}(\mathsf{Alice}))
 \end{aligned}
\end{equation*}
\end{footnotesize}
\vspace{-0.10in}

\noindent In order to flip the last two statements above, we need to
modify $\Gamma$, but we can derive that $\mathsf{Alice}$ is the
murderer without any modifications, and since
$\delta^a_t(\Gamma, \Gamma)= 0$, it easier to believe $\mathsf{Alice}$ is
the murderer than to withhold that $\mathsf{Alice}$ is
the murderer. $\blacksquare$

\vspace{-0.06in}

\subsection{Counterfactuals}
At time $t$, assume that an agent $a$ believes in a set of
propositions $\Gamma$ and is interested in propositions
$\holds(f, t')$ and $\holds(g, t')$ with $t'<t$ and:
$$\Gamma\vdash \lnot \holds(f,t') \land \lnot \holds(g,t')$$
We may need non-trivial uncertainty values, but in this case,
$\mathbf{Pr}$ will assign a trivial value of $0$ to both the
propositions. We can then look at closest consistent sets to $\Gamma$
under $\delta$:
\begin{footnotesize}
\begin{equation*}
  \begin{aligned}
\Gamma_1 &\vdash \holds(f, t') \\
\Gamma_2 &\vdash \holds(g, t') 
 \end{aligned}
\end{equation*}
\end{footnotesize}
Clause III from the definition for reasonableness gives us: 
\vspace{-0.06in}
\begin{footnotesize}
\begin{equation*}
  \begin{aligned}
    \believes\Big(a, t,  \holds(f, t') \Big) &\succ^a_t
    \believes\Big(a, t,  \holds(g, t') \Big) \\
&\Leftrightarrow\\
\delta^a_t(\Gamma, \Gamma_1) &< \delta^a_t(\Gamma, \Gamma_2) 
 \end{aligned}
\end{equation*}
\end{footnotesize}

\vspace{-0.1in}

\subsection{Explanations}
The definitions of the strength factors and reasonableness above can
be used to generate high-level schemas for explanations. These schemas
can be used instead of simply presenting raw probability values to
end-users. While we have not fleshed out such explanation schemas, we
illustrate one possible schema. Say an agent performs an action
$\alpha$ on the basis of $\phi$. In this case, the agent could
generate an explanation that at the highest level simply says that it
is more reasonable for the agent to believe $\phi$ than for the agent
to believe in $\lnot \phi$. The agent could then further explain why
it was reasonable for it by using one of the three clauses in the
reasonableness definition.

\vspace{-0.06in}


\section{Inference Algorithm Sketch}

Describing the inference algorithm in detail is beyond the scope of
this paper, but we provide a high-level sketch here.\footnote{More
  details can be found here: \url{https://goo.gl/2Vz2nJ}} Our proof
calculus is simply an extension of standard first-order proof calculus
under different modal contexts. For example, if $a$ believes that $b$
believes in a set of propositions $\Gamma$ and
$\Gamma\vdash_{FOL} \psi$, then $a$ believes that $b$ believes $\psi$.
We convert $\believes\left(a, t_a, \believes\left(b, t_b, Q\right)\right)$
into the pure first-order formula $Q\left(context(a, t_a, b, t_b)\right)$
and use a first-order prover. The conversion process is a bit more
nuanced as we have to handle negations, properly handle substitutions
of equalities, uncertainties and transform compound formulae within
iterated beliefs.

\vspace{-0.06in}

\section{Conclusion and Future Work}
We have presented initial steps in building a system of uncertainty
that is both probability and proof theory based that could lend itself
to \begin{inparaenum}[(1)] \item solving foundational problems; \item being
  useful in applications; \item generating uncertainty values for
  counterfactuals; and \item building understandable
  explanations. \end{inparaenum}

Shortcomings of $\mathcal{S}$ can be cast as challenges, and many
challenges exist, some relatively easy and some quite hard. Among the
easy challenges are defining and experimenting with different
candidates for $\mathbf{Pr}$, $\rho$, $\pi$ and $\delta$. On the more
difficult side, we have to come up with tractable computational
mechanisms for computing the $\min_{\langle \Theta, \Lambda \rangle}$
in the definition for $\delta$. Also on the difficult side, is the
challenge of coming up efficient reasoning schemes. While we have an
exact inference algorithm, we believe that an approximate algorithm
that selectively discards beliefs in a large knowledge base during
reasoning will be more useful. Future work also includes comparison
with other uncertainty systems and exploration of conditions under
which uncertainty values of $\mathcal{S}$ are similar/dissimilar with
other systems (thresholded appropriately).

\vspace{-0.06in}

\section*{Acknowledgements}
\label{sect:ack}

We are grateful to the Office of Naval Research for their funding of
projects titled \emph{``Advanced Logicist Machine Learning''} and
\emph{``Making Morally Competent Robots''} and to the Air Force Office
of Scientific Research for funding the project titled \emph{``Great
  Computation Intelligence: Mature and Further Applied''} that enabled
the research presented in this paper.  We are also thankful for the
insightful reviews provided by the three anonymous referees.


\appendix

\vspace{-0.05in}

\section{Appendix: Slate Proofs}
\label{sect:appendixA}

The figures below are vector graphics and can be zoomed to more easily
read the contents. 
\vspace{-0.20in}

\begin{figure}[h!]
 \begin{minipage}[b]{\linewidth}
\caption{$\mathsf{Alice}$ is the murder: $\believes\Big(s, t,  \mathit{Murderer}(\mathsf{Alice})\Big)$}  \medskip
  \label{fig:proof1}
  {\includegraphics[scale=0.35]{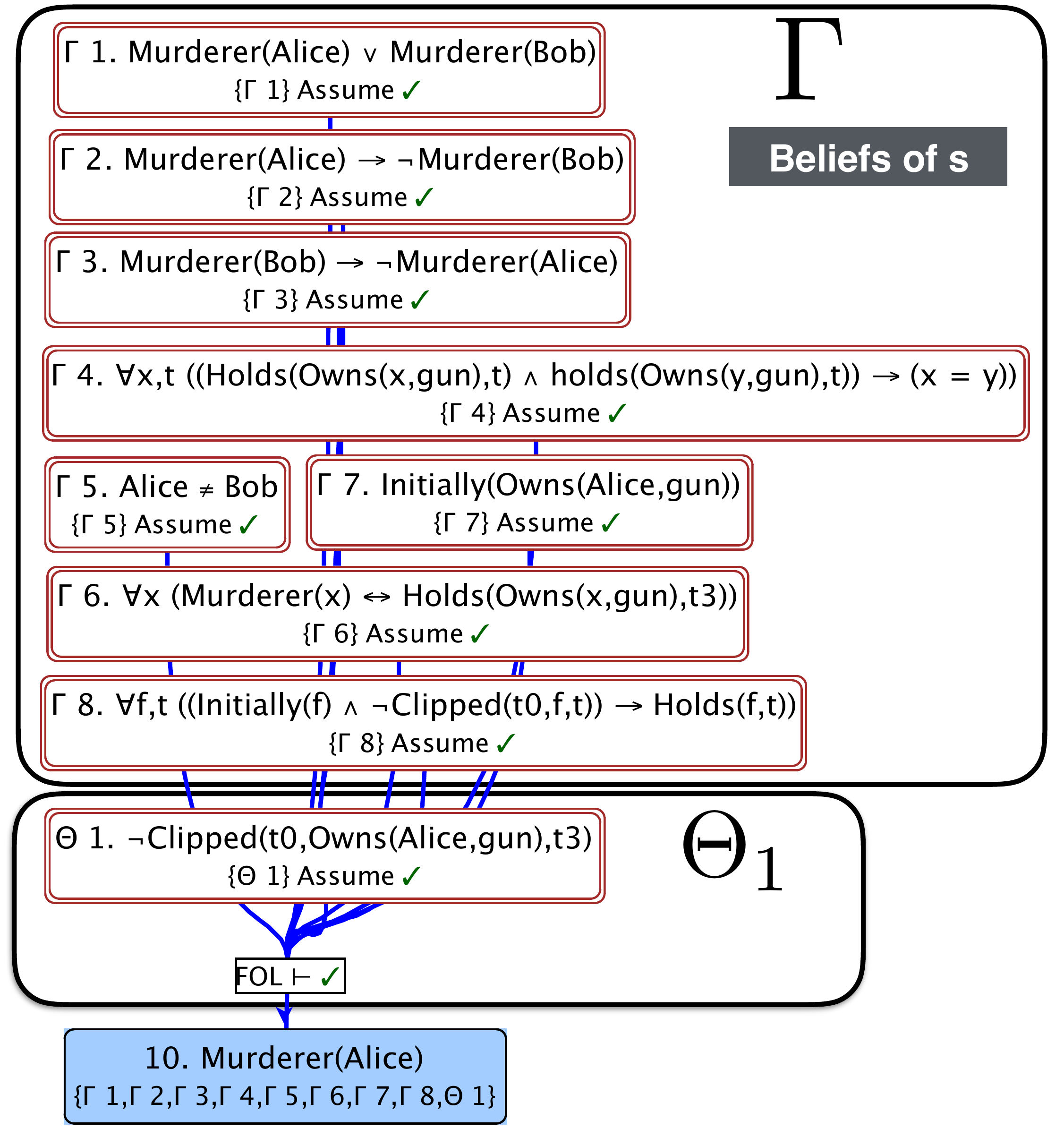}}
 \end{minipage}
\end{figure}
\vspace{-0.3in}

\clearpage \newpage
\begin{small}
\bibliographystyle{named}
\bibliography{main72,naveen}
\end{small}

\end{document}

%% file: lottery_paradox.tex
Let $\Gamma_{L}$ be a meticulous and perfectly accurate description of a
1,000,000,000,000-ticket lottery, of which rational agent $a$ is fully
apprised. Assume that from $\Gamma_{L}$  it can be proved that either
ticket 1 will win or ticket 2 will win or $\ldots$ or ticket
1,000,000,000,000 will win.  Let’s write this (exclusive) disjunction
as follows (here $\oplus$ is an exclusive disjunction):

\begin{footnotesize}
\begin{equation*}
  \begin{aligned}
\Gamma_{L} \vdash win(t_1) \oplus win(t_2) \oplus \ldots \oplus win(t_{1,000,000,000,000})  
  \end{aligned}
\end{equation*}
\end{footnotesize}

The paradox has two strands of reasoning. The first strand yields
$\believes(a, now, \phi)$ and the second strand yields
$\believes(a, now, \lnot\phi)$ with
$\phi \equiv \exists t: win(t)$. 

\noindent \textbf{Strand 1}: Since $a$ believes all propositions in
$\Gamma_L$, $a$ can then deduce from this the belief that there is at
least one ticket that will win, a proposition represented as:

\begin{footnotesize}
\begin{equation*}
  \begin{aligned}
\fbox{S$_1$} \hspace{10pt} \mathbf{B}\Big(a, now, \exists t: win\big(t\big)\Big) 
  \end{aligned}
\end{equation*}
\end{footnotesize}

\noindent \textbf{Strand 2}: From $\Gamma_L$ it can be proved that the
probability of a particular ticket $t_i$ winning is $10^{-12}$. 

\begin{footnotesize}
\begin{equation*}
  \begin{aligned}
\Big[\mathbf{Pr}\Big(a, now, win\big(t_1\big)\Big) = 10^{-12}\Big]
\wedge &\Big[\mathbf{Pr}\Big(a, now, win\big(t_2\big)\Big)=
10^{-12}\Big]\\ \wedge
\ldots \wedge  &\Big[\mathbf{Pr}\Big(a, now, win\big(t_{1T}\big)\Big)= 10^{-12}\Big]
  \end{aligned}
\end{equation*}
\end{footnotesize}

For the next step, note that the probability of ticket $t_1$ winning
is lower than, say, the probability that if you walk outside a minute
from now, you will be flattened on the spot by a door from a 747 that
falls from a jet of that type cruising at 35,000 feet. Since you, the
reader, have the rational belief that death won't ensue if you go
outside (and have this belief precisely because you believe that the
odds of your sudden demise in this manner are vanishingly small), the
inference to the rational belief on the part of $a$ that $t_1$ won't
win sails through --- and this of course works for each ticket. Hence
we have as a valid belief (though not derivable in \mC\ from
$\Gamma_L$):

\begin{footnotesize}
\begin{equation*}
  \begin{aligned}
&\mathbf{B}\Big(a, now, \neg win\big(t_1)\Big) \wedge \mathbf{B}\Big(a, now, \neg
win\big(t_2)\Big)\wedge \ldots \\ 
&\wedge \mathbf{B}\Big(a, now, \neg win\big(t_{1T})\Big)
  \end{aligned}
\end{equation*}
\end{footnotesize}

From $R_{\mathbf{B}}$ and above, we get:

\begin{footnotesize}
\begin{equation*}
  \begin{aligned}
\mathbf{B}\Big(a,now,  \neg win\big(t_1) \land  \neg win\big(t_2)\land
\ldots \land  \neg win\big(t_{1T})\Big)  \end{aligned}
\end{equation*}
\end{footnotesize}

Applying $R_{\mathbf{B}}$ to the above and $\Gamma_L$, we get:
\begin{footnotesize}
\begin{equation*}
  \begin{aligned}
\fbox{S$_2$} \hspace{10pt}\mathbf{B}\Big(a, now, \neg \exists t: win(t)\Big) 
  \end{aligned}
\end{equation*}
\end{footnotesize} 

The two strands are complete, and we have derived contradictory
beliefs labeled S$_1$ and S$_2$. Our solution consists of two new
uncertainty infused strands that result in beliefs of sufficiently
varying strengths that block inferences that could combine them.

\noindent \textbf{Strand 3}: Assume that $a$ is certain of all
propositions in $\Gamma_L$, then using $R^s_{\believes}$, we have: 

\begin{footnotesize}
\begin{equation*}
  \begin{aligned}
\fbox{S$_3$} \hspace{10pt} \mathbf{B}^5\big(a, now, \exists t: win(t)\big) 
  \end{aligned}
\end{equation*}
\end{footnotesize}

\noindent \textbf{Strand 4}: Since $\mathbf{Pr} (a, now, win(t_i)) <
\mathbf{Pr}(a, now, \lnot  win(t_i))$, using Clause I and the strength
factor definitions,  we have now that for all $t_i$

\begin{footnotesize}
\begin{equation*}
  \begin{aligned}
\mathbf{B}^2\big(a, now,  \lnot win(t_i)\big) 
  \end{aligned}
\end{equation*}
\end{footnotesize}

Using the reasoning similar to that in Strand 2, we get:

\begin{footnotesize}
\begin{equation*}
  \begin{aligned}
\fbox{S$_4$} \hspace{10pt}\mathbf{B}^2\Big(a, now, \neg \exists t:
win(t)\Big) 
  \end{aligned}
\end{equation*}
\end{footnotesize} 

Strands 3 and 4 resolve the paradox. Note that $R^s_{\believes}$
cannot be applied to S$_3$ and S$_4$ and churn out arbitrary
propositions, as the default value of the $u$ parameter in
$R^s_{\believes}$ requires beliefs to be no more than 2 levels apart. $\blacksquare$


%% file: main.bbl
\begin{thebibliography}{}

\bibitem[\protect\citeauthoryear{Arkoudas and
  Bringsjord}{2008}]{ArkoudasAndBringsjord2008Pricai}
Konstantine Arkoudas and Selmer Bringsjord.
\newblock {Toward Formalizing Common-Sense Psychology: {A}n Analysis of the
  False-Belief Task}.
\newblock In T.-B. Ho and Z.-H. Zhou, editors, {\em {Proceedings of the Tenth
  Pacific Rim International Conference on Artificial Intelligence (PRICAI
  2008)}}, number 5351 in Lecture Notes in Artificial Intelligence (LNAI),
  pages 17--29. Springer-Verlag, 2008.

\bibitem[\protect\citeauthoryear{Bringsjord \bgroup \em et al.\egroup
  }{2008}]{Slate_at_CMNA08}
Selmer Bringsjord, Joshua Taylor, Andrew Shilliday, Micah Clark, and
  Konstantine Arkoudas.
\newblock {Slate: An Argument-Centered Intelligent Assistant to Human
  Reasoners}.
\newblock In Floriana Grasso, Nancy Green, Rodger Kibble, and Chris Reed,
  editors, {\em {Proceedings of the 8th International Workshop on Computational
  Models of Natural Argument (CMNA 8)}}, pages 1--10, Patras, Greece, July 21
  2008. University of Patras.

\bibitem[\protect\citeauthoryear{Bringsjord \bgroup \em et al.\egroup
  }{2014}]{akratic_robots_ieee_n}
Selmer Bringsjord, Naveen~Sundar Govindarajulu, Daniel Thero, and Mei Si.
\newblock {Akratic Robots and the Computational Logic Thereof}.
\newblock In {\em Proceedings of {\textit{ETHICS} $\bullet$ 2014} (2014 IEEE
  Symposium on Ethics in Engineering, Science, and Technology)}, pages 22--29,
  Chicago, IL, 2014.
\newblock {IEEE Catalog Number: CFP14ETI-POD.}

\bibitem[\protect\citeauthoryear{Chisholm}{1987}]{theory.of.knowledge3.chisholm}
Roderick Chisholm.
\newblock {\em Theory of Knowledge 3rd ed}.
\newblock Prentice-Hall, Englewood Cliffs, NJ, 1987.

\bibitem[\protect\citeauthoryear{Demey \bgroup \em et al.\egroup
  }{2016}]{sep-logic-probability}
Lorenz Demey, Barteld Kooi, and Joshua Sack.
\newblock {Logic and Probability}.
\newblock In Edward~N. Zalta, editor, {\em The Stanford Encyclopedia of
  Philosophy}. Metaphysics Research Lab, Stanford University, winter 2016
  edition, 2016.

\bibitem[\protect\citeauthoryear{Gentzen}{1935}]{gentzen_investigations_into_logical_deduction}
Gerhard Gentzen.
\newblock {Investigations into Logical Deduction}.
\newblock In M.~E. Szabo, editor, {\em The Collected Papers of Gerhard
  Gentzen}, pages 68--131. North-Holland, Amsterdam, The Netherlands, 1935.
\newblock {This is an English version of the well-known 1935 German version.}

\bibitem[\protect\citeauthoryear{Govindarajulu and
  Bringsjord}{2017}]{nsg_sb_dde_2017}
Naveen~Sundar Govindarajulu and Selmer Bringsjord.
\newblock {On Automating the Doctrine of Double Effect}.
\newblock In {\em {Proceedings of the Twenty-Sixth International Joint
  Conference on Artificial Intelligence ({IJCAI} 2017)}}, 2017.
\newblock Preprint available at this url:
  \url{https://arxiv.org/abs/1703.08922}.

\bibitem[\protect\citeauthoryear{Halpern}{1990}]{halpern1990analysis}
Joseph~Y Halpern.
\newblock {An Analysis of First-order Logics of Probability}.
\newblock {\em Artificial intelligence}, 46(3):311--350, 1990.

\bibitem[\protect\citeauthoryear{Kaye and Koehler}{1991}]{kaye_1991}
D.~H. Kaye and Jonathan~J. Koehler.
\newblock {Can Jurors Understand Probabilistic Evidence?}
\newblock {\em Journal of the Royal Statistical Society. Series A (Statistics
  in Society)}, 154(1):75--81, 1991.

\bibitem[\protect\citeauthoryear{Kyburg~Jr}{1961}]{kyburg1961probability}
Henry~E Kyburg~Jr.
\newblock {\em {Probability and the Logic of Rational Belief}}.
\newblock Wesleyan University Press, Middletown, CT, 1961.

\bibitem[\protect\citeauthoryear{Licato \bgroup \em et al.\egroup
  }{2013}]{GIfromLP_ijcai13_short}
John Licato, Naveen~Sundar Govindarajulu, Selmer Bringsjord, Michael Pomeranz,
  and Logan Gittelson.
\newblock {Analogico-Deductive Generation of G\"{o}del's First Incompleteness
  Theorem from the Liar Paradox}.
\newblock In Francesca Rossi, editor, {\em Proceedings of the 23rd
  International Joint Conference on Artificial Intelligence ({IJCAI--}13)},
  pages 1004--1009, Beijing, China, 2013. Morgan Kaufmann.

\bibitem[\protect\citeauthoryear{Nunez \bgroup \em et al.\egroup
  }{2013}]{nunezetal13folinds}
Rafael~C. Nunez, Matthias Scheutz, Kamal Premaratne, and Manohar~N. Murthi.
\newblock {Modeling Uncertainty in First-Order Logic: A Dempster-Shafer
  Theoretic Approach}.
\newblock In {\em 8th International Symposium on Imprecise Probability:
  Theories and Applications}, 2013.

\bibitem[\protect\citeauthoryear{Zalta}{1988}]{zalta1988intensional}
Edward~N Zalta.
\newblock {\em {Intensional Logic and the Metaphysics of Intentionality}}.
\newblock MIT Press, Cambridge, MA, 1988.

\end{thebibliography}
